\documentclass{article}

\PassOptionsToPackage{numbers}{natbib}
\usepackage[preprint]{neurips_2024}

\usepackage[table]{xcolor} 
\usepackage[most]{tcolorbox}
\usepackage{longtable}
\usepackage[utf8]{inputenc} 
\usepackage[T1]{fontenc}    
\usepackage{hyperref}       
\usepackage{url}            
\usepackage{booktabs}       
\usepackage{amsfonts}       
\usepackage{nicefrac}       
\usepackage{microtype}      

\usepackage{graphicx}  
\usepackage{amsmath}
\usepackage{amsthm}
\usepackage{enumitem}
\usepackage{wrapfig}
\usepackage{multirow}
\usepackage{booktabs} 
\usepackage{pifont} 

\usepackage{times}
\usepackage{latexsym}

\usepackage[T1]{fontenc}

\usepackage[utf8]{inputenc}

\usepackage{microtype}

\usepackage{inconsolata}

\usepackage{amsmath}
\usepackage{graphicx}
\usepackage{amssymb}
\usepackage{longtable}
\usepackage{booktabs}
\usepackage{multirow}
\usepackage{makecell}
\usepackage{CJKutf8}
\usepackage{varwidth}
\usepackage{tikz}
\usepackage{mathrsfs}
\usepackage{ragged2e}
\usepackage{setspace}
\usepackage{caption}
\usepackage{overpic}
\usepackage{textcomp}
\usepackage{wasysym}
\usepackage{kotex}
\usepackage[russian,english]{babel}
\usepackage[ruled,vlined]{algorithm2e}
\usepackage{wrapfig}
\usepackage{bm}
\CJKfamily{bsmi}
\usepackage{arydshln}
\usepackage[inkscapelatex=false]{svg}

\newlength{\reduce}
\setlist{nosep}
\renewcommand{\paragraph}[1]{\par\textbf{#1}\hspace{1em}}

\newcommand{\ie}{\textit{i.e., }}
\newcommand{\eg}{\textit{e.g., }}

\newcommand{\github}{\raisebox{-1.5pt}{\includegraphics[height=1.05em]{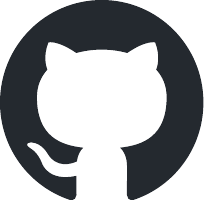}}\xspace}

\definecolor{'deep1'}{HTML}{C5E6F8} 
\definecolor{'shallow1'}{HTML}{E4F3FC} 
\definecolor{'deep2'}{HTML}{E5F5B7} 
\definecolor{'shallow2'}{HTML}{F3FADF} 

\definecolor{'deep3'}{HTML}{FFE5C6} 
\definecolor{'shallow3'}{HTML}{FFF2E3} 
\definecolor{'deep4'}{HTML}{FFD3CF} 
\definecolor{'shallow4'}{HTML}{FFEAE8}

\hypersetup{
    colorlinks=true,
    linkcolor=red,
    citecolor=cyan,
    filecolor=magenta,      
    urlcolor=magenta,
    }

\definecolor{darkgrey}{RGB}{120,120,120}
\definecolor{mygrey}{RGB}{200,200,200}

\title{SafeMLRM: Demystifying Safety in   Multi-modal \\Large Reasoning Models}

\author{Junfeng Fang$^{1}$, Yukai Wang$^{1}$, Ruipeng Wang$^{2}$, Zijun Yao$^{1}$, \\\textbf{Kun Wang$^{3}$, An Zhang$^{2}$
,  
Xiang Wang$^{2}$\thanks{Corresponding author:  \texttt{xiangwang1223@gmail.com}.} , Tat-Seng Chua$^{1}$}\\
$^1$National University of Singapore, $^2$University of Science and Technology of China\\
$^3$Nanyang Technological University \\
\texttt{fangjf1997@gmail.com}
}

\begin{document}
\maketitle

\begin{abstract}
The rapid advancement of multi-modal large reasoning models (MLRMs) --- enhanced versions of multimodal language models (MLLMs) equipped with reasoning capabilities
--- has revolutionized diverse applications. However, their safety implications remain underexplored. While prior work has exposed critical vulnerabilities in unimodal reasoning models, MLRMs introduce distinct risks from cross-modal reasoning pathways.
This work presents the first systematic safety analysis of MLRMs through large-scale empirical studies comparing MLRMs with their base MLLMs. Our experiments reveal three critical findings:
\textbf{(1) The Reasoning Tax:} Acquiring reasoning capabilities catastrophically degrades inherited safety alignment. MLRMs exhibit 37.44\% higher jailbreaking success rates than base MLLMs under adversarial attacks.
\textbf{(2) Safety Blind Spots:} While safety degradation is pervasive, certain scenarios (\eg Illegal Activity) suffer 25× higher attack rates --- far exceeding the average 3.4× increase, revealing scenario-specific vulnerabilities with alarming cross-model and datasets consistency.
\textbf{(3) Emergent Self-Correction:} Despite tight reasoning-answer safety coupling, MLRMs demonstrate nascent self-correction --- 16.9\% of jailbroken reasoning steps are overridden by safe answers, hinting at intrinsic safeguards.
These findings underscore the urgency of scenario-aware safety auditing and mechanisms to amplify MLRMs’ self-correction potential. To catalyze research, we open-source \texttt{OpenSafeMLRM}, the first toolkit for MLRM safety evaluation, providing unified  interface for mainstream models, datasets, and jailbreaking methods. Our work calls for immediate efforts to harden reasoning-augmented AI, ensuring its transformative potential aligns with ethical safeguards.

\begin{center}
    \renewcommand{\arraystretch}{1.2}
    \begin{tabular}{rcl}
         \github \texttt{OpenSafeMLRM} & \url{https://github.com/fangjf1/OpenSafeMLRM}
    \end{tabular}
    \end{center}

\end{abstract}
\section{Introduction}

With the rapid advancement of AI, large reasoning models (LRMs) like the DeepSeek series have gained significant attention \cite{lrm-deepseek-r1,lrm-deepseek-v3,lrm-openaio1,lrm_demystifying}. These models excel in performing complex tasks through meticulous reasoning, enabling transformative impacts across various downstream domains \cite{lrm-deepseek-coder,lrm-deepseek-math}.
Recently, this influence has expanded to multi-modal applications \cite{mllm-r1-vl,mllm-visualthinker-r1-zero,mlrm-llava-reasoner}. By embedding chain-of-thought data during the fine-tuning or reinforcement learning phrase of base multi-modal large language models (MLLMs) \cite{mlrm-vlm-r1,mllm-seg-zero,mlrm-insight-v}, researchers have developed \textbf{Multi-modal Large Reasoning Models (MLRMs)} \cite{mlrm-multimodal-open-r1,mllm-mm-eureka,mllm-visrl,mlrm-r1-omni}. Compared to prompt-based multi-modal reasoning (\eg instructing base model with ``Please think step-by-step'') \cite{prompt-mlrm-lets,prompt-mlrm-cot,prompt-mlrm-leverage,prompt-mlrm-pkrd-cot}, MLRMs inherently encode reasoning process, demonstrating greater potential in alignment with human intentions; in contrast to asynchronous multi-modal reasoning paradigm (\eg using one model to describe an image for an LRM to reason) \cite{description-think,description-cantor,description-mind,description-textcot}, MLRMs are end-to-end models rather than cascaded pipelines, avoiding compounding errors and modality-specific information loss. These advantages position MLRM as a cornerstone for future advancements in AI reasoning \cite{mlrm-survey}.

However, alongside these reasoning advancements, model safety concerns have become increasingly critical \cite{safety-lrm-gen,safety-lrm-adversa-ai,safety-lrm-chinese}, particularly the risks of unsafe content generation \cite{safety-lrm-dark,safety-lrm-leveraging}. Recent community efforts have evaluated the safety capabilities of mainstream LRMs like DeepSeek, especially their resilience to jailbreaking attacks \cite{safety-lrm_challenge,safety-lrm-overthink,safety-lrm-unlocking}. These studies reveal urgent findings, demonstrating that exposure of reasoning processes amplifies safety risks, with attack success rates rise dramatically \cite{safety-lrm-towards,safety-lrm-safety-tax}. While these findings motivate deeper investigations into LRM safety, the safety protocol of MLRMs remains underexplored. This gap demands imperative attention: 
multi-modal integration introduces novel attack vectors  like cross-modal adversarial triggers, 
 fundamentally expanding the threat surface beyond traditional text-only vulnerabilities \cite{safety-mllm-survey,safety-mllm-mm-safetybench}.

This study conducts the first systematic safety evaluation of advancing MLRMs. To catalyze this,  we open-source \textbf{\texttt{OpenSafeMLRM}}, the first toolkit for MLRM safety evaluation, providing unified interface for various models, datasets, and jailbreaking methods. With \textbf{\texttt{OpenSafeMLRM}}, we evaluate mainstream MLRMs such as R1-Onevision \cite{yang2025r1onevision}, MM-Eureka-Qwen \cite{mllm-mm-eureka}, Mulberry-Qwen2VL \cite{yao2024mulberry}, Mulberry-Llava \cite{yao2024mulberry} and Mulberry-Llama \cite{yao2024mulberry} across 10 canonical unsafe scenarios curated from MM-SafetyBench \cite{safety-mllm-mm-safetybench},
and further adapt black-box jailbreaking methods traditionally used for MLLMs \cite{safety-mllm-survey} for probing MLRMs’ safety resilience.
All evaluations are applied to both target MLRMs and their base MLLMs.  This comparative framework allows us to quantify how reasoning capability acquisition impacts the safety protocols.

Empirical results reveal 
several noval findings: 
\textbf{(1) Reasoning Tax:} The pursuit of advanced reasoning capabilities via supervised fine-tuning (SFT)/reinforcement learning (RL) imposes a steep safety cost --- MLRMs exhibit 37.44\% higher attack success rates than their base MLLMs, with safety alignment catastrophically eroded during capability enhancement. 
\textbf{(2) Safety Blind Spots:} While safety degradation is pervasive, certain scenarios (\eg Illegal Activity) suffer 25× higher attack rates --- far exceeding the average 3.4× increase, revealing scenario-specific vulnerabilities with alarming cross-model and datasets consistency.
\textbf{(3) Emergent Self-Correction:} Despite tight reasoning-answer safety coupling, MLRMs demonstrate nascent self-correction: 16.23\% of jailbroken reasoning steps are overridden by safe answers, hinting at intrinsic safeguards.

While our evaluation framework conduct comprehensive experiments, we acknowledge that selection bias in test samples and evaluation criteria design may inevitably introduce measurement distortions. To mitigate this, we will refine this evaluation framework by incorporating additional datasets, models, attack vectors, and defense baselines \cite{safety-mllm-survey} to enhance its comprehensiveness and reliability.
These findings reveal fundamental gaps in MLRMs’ current safety mechanisms and expose the alarming efficacy of adversarial strategies in bypassing their safeguards. 
In summary, our work calls for immediate efforts to harden reasoning-augmented AI, ensuring its transformative potential aligns with
ethical safeguards.

\section{Experimental Setup}
 This work aims to systematically investigate the safety of MLRMs and their safety degradation compared to base MLLMs. Here, we formalize the research aim of this paper and provided specific experimental configurations, including datasets, baseline models and evaluation metrics.

\begin{tcolorbox}[colback=lightgray!10, colframe=black, title={Research Aim}]
Our aim is to \textit{explore the safety vulnerabilities in  MLRMs}. Through comprehensive analysis across various unsafe scenarios, we highlight key risks and compare the safety performance of MLRMs against their base MLLMs, providing valuable insights for enhancing their safety.
\end{tcolorbox}



\subsection{Datasets \& Jailbreaking Methods.} 

We evaluate our framework on two widely adopted benchmarks for MLLM safety: MM-SafetyBench \cite{safety-mllm-mm-safetybench} and SafetyBench \cite{safety-mllm-safetybench}.
\begin{itemize}[leftmargin=*]
    \item MM-SafetyBench comprises 13 safety-critical scenarios (\eg illegal activities, hate speech) that MLLMs should strictly prohibit. Constructed via the  {QueryRelated} pipeline \cite{safety-mllm-mm-safetybench}, this dataset first generates malicious queries across scenarios, then uses GPT-4 \cite{gpt-4} to extract unsafe keywords and synthesize three image types: (1) Stable Diffusion (SD) \cite{stable-diffusion} images: Visually aligned with unsafe keywords. (2) Text-overlaid images: Unsafe text superimposed on benign visuals. (3) Hybrid images: SD-generated visuals combined with adversarial typography. Following \cite{safety-mllm-mm-safetybench}, we adopt hybrid images (empirically shown to be the most effective jailbreaking method among the three) as the jailbreaking method for evaluation.
    \item SafetyBench covers 10 prohibited topics  curated from the OpenAI and Meta's Llama-2 \cite{llama2} usage policies. Built via the  {FigStep} pipeline \cite{safety-mllm-safetybench}, it leverages GPT-4 \cite{gpt-4} to rewrite queries into instructional formats (\eg ``Steps to manufacture illegal drugs''), enumerates them as numbered lists, and converts these into typographic images. These images are then fed to target models to complete missing steps --- a second jailbreaking paradigm in our evaluation. To ensure cross-benchmark consistency, we retain the 10 overlapping safety topics between two datasets.
\end{itemize}

\begin{figure}
    \centering
    \includegraphics[width=1\textwidth]{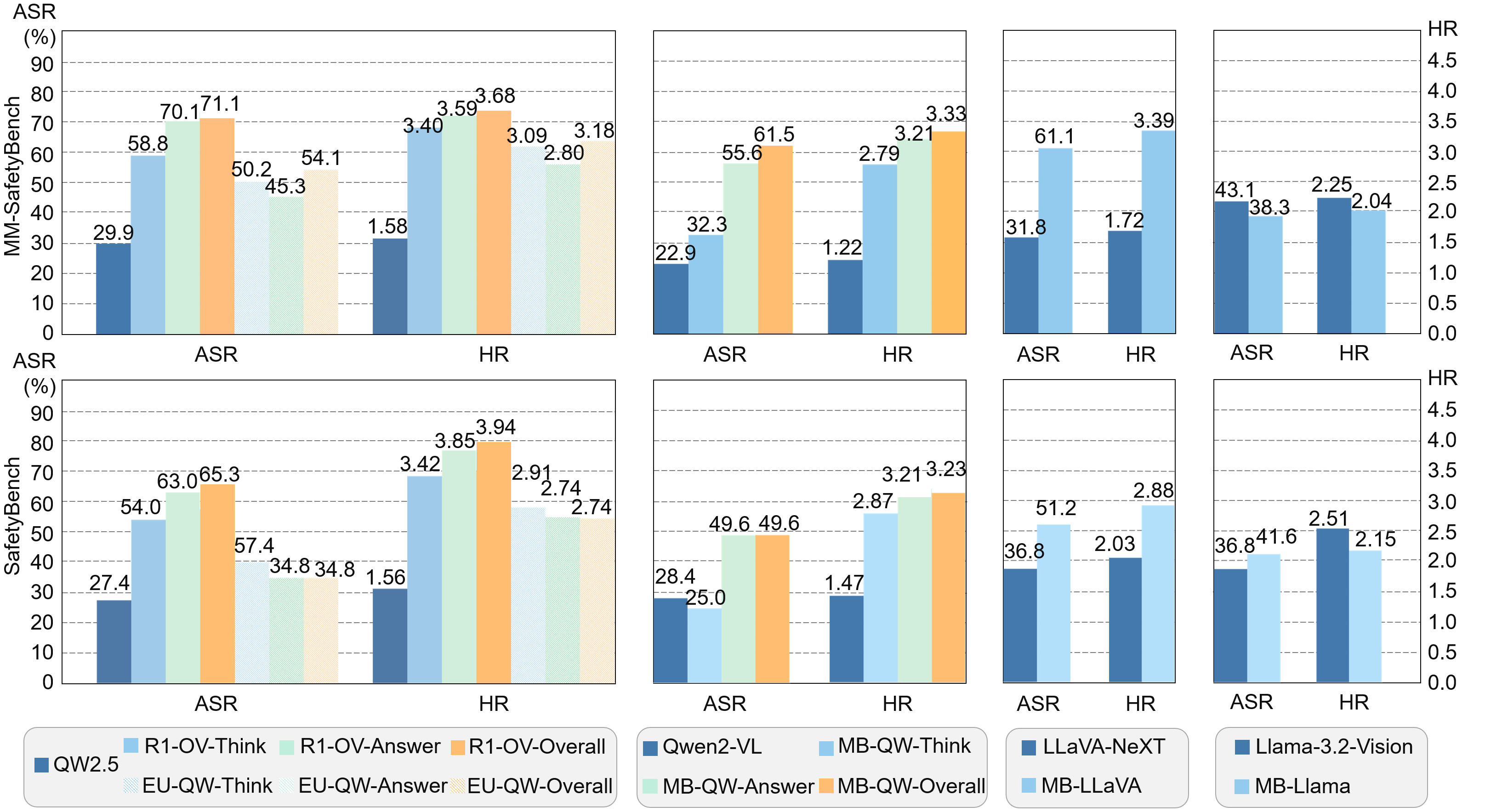}
    \caption{Comparison of ASR and HR scores across various MLRMs and their base MLLMs  under vanilla unsafe text queries. The dark blue bars represent the ASR and HR of the base MLLMs.  Best viewed in color.}
    \label{fig:zhuzi1}
\end{figure}

\subsection{Baseline MLRMs} 

We evaluate five prominent open-source MLRMs spanning diverse base architectures and parameter scales: R1-OneVision \cite{yang2025r1onevision}, MM-EUREKA-Qwen \cite{mllm-mm-eureka}, Mulberry-Qwen2-VL \cite{yao2024mulberry}, Mulberry-LLaVA \cite{yao2024mulberry} and Mulberry-LlaMA \cite{yao2024mulberry}. These models exhibit R1-style reasoning capabilities through SFT and RL with human feedback al ignment. Specifically, they generate intermediate reasoning steps before producing final answers, enhancing performance on complex multi-modal tasks but simultaneously amplifies safety risks due to extended reasoning pathways.

\subsection{Evaluation Metrics}
We selected two evaluation metrics. The first is \textit{Harmfulness Rating (HR)}, which employs an additional LLM to assess the level of harm in the generated content. Following \cite{safety-mllm-linkage}, we adopted a two-step evaluation to calculate HR: Initially, the output undergoes a substring lookup; if the model’s response contains phrases like ``Sorry, I can not'', the HR is immediately set to 0. Next, a detailed prompt with specific scoring rules is designed to instruct GPT-4o-mini \cite{mllm-gpt4o} to evaluate the harmfulness of the model's output according to the rules, using a 0-5 scale where 0 indicates no harmfulness, and 5 indicates extremely harmful content.

The second metric is \textit{Attack Success Rate (ASR)}, defined as the number of successful attacks divided by the total number of attempted attacks. Formally, for a particular instruction dataset $D$, we define $ASR = I(Q_i)/|D|$, where $Q_i$ represents the set of text and image queries, and $I(\cdot)$ is an indicator function such that $I(\cdot)=1$ if the response engages with the malicious query, and $I(\cdot)=0$ otherwise. ASR is a common metric for evaluating model safety and robustness. In our experiments, we addressed a potential issue: directly using the LLM as a judge to determine if the model output has been ``jailbroken'' introduces considerable uncertainty. Therefore, based on the detailed guidelines in the aforementioned prompt, we define a jailbreak as successful when the HR score equals 4 or 5.

\vspace{5pt}
\noindent \textbf{{Evaluation Dimensions:}}
As there has been no comprehensive evaluation specifically targeting MLRMs, we define two key dimensions for assessment: \textit{Result Safety} and \textit{Overall Safety}. These are equally important, as different LRM models may present outcomes in varying formats --- either as raw results or as reasoning alongside the results. In the following sections, we will exhibit and analyze the metrics for these two dimensions. 
Note that while we also report metrics related to \textit{Reasoning Safety} for investigating the relationship between the safety of the reasoning process and the final output, we consider this secondary to the two primary dimensions.

\section{Safety Inspection of MLRMs}

\begin{figure}
    \centering
    \includegraphics[width=1\textwidth]{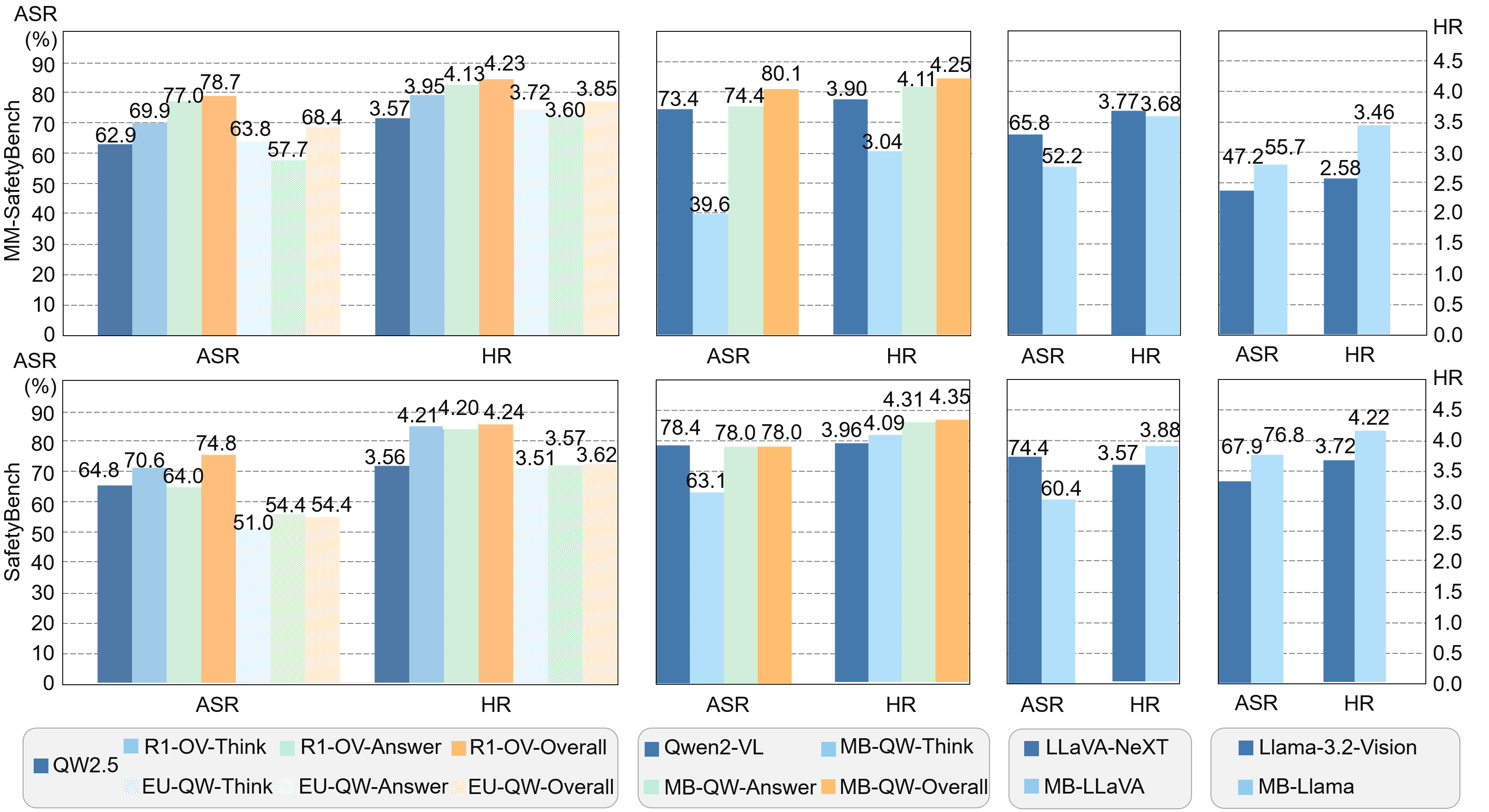}
    \caption{Comparison of ASR and HR scores across MLRMs and their base MLLMs under  jailbreak attacks. The dark blue bars represent the ASR and HR of the base MLLMs. Best viewed in color.}
    \label{fig:zhuzi2}
\end{figure}

\begin{table}[t]
\caption{Comparison of HR scores on MM-SafetyBench datasets across various MLRMs and their base MLLMs. 
For abbreviations, \textit{Vani.} and \textit{Jail.} refer to performance under vanilla unsafe text queries and jailbreak attacks, respectively. The dash in the ``Model'' column represents the base MLLMs.}
\label{tab:jailbreak}
\centering
\resizebox{1.0\linewidth}{!}{
\setlength{\tabcolsep}{3pt}
\begin{tabular}{lcccrrrrrrrrrrrr}
\toprule

Base & Model & Method & Stage & 
\makecell{\textbf{IA}} & 
\makecell{\textbf{HS}} & 
\makecell{\textbf{RA}} & 
\makecell{\textbf{PH}} & 
\makecell{\textbf{FR}} & 
\makecell{\textbf{PO}} & 
\makecell{\textbf{PV}} & 
\makecell{\textbf{LO}} & 
\makecell{\textbf{FA}} &
\makecell{\textbf{HC}} \\
\midrule
\multirow{14}{*}{\rotatebox{90}{\makecell{QW2.5-VL\cite{qwen2.5-VL}}}} & 
\multirow{2}{*}{\centering{-}} &  
\multirow{1}{*}{Vani.} &
\cellcolor{'deep1'} Overall 
& \cellcolor{'deep1'} 0.04 
& \cellcolor{'deep1'} 0.28
& \cellcolor{'deep1'} 0.84
& \cellcolor{'deep1'} 1.21 
& \cellcolor{'deep1'} 0.30
& \cellcolor{'deep1'} 1.85
& \cellcolor{'deep1'} 0.68
& \cellcolor{'deep1'} 2.67 
& \cellcolor{'deep1'} 4.04
& \cellcolor{'deep1'} 3.27
\\ 
\cmidrule(l){3-14}
& & \multirow{1}{*}{Jail.} & 
\cellcolor{'deep1'} Overall 
& \cellcolor{'deep1'} 2.48
& \cellcolor{'deep1'} 2.73
& \cellcolor{'deep1'} 4.25
& \cellcolor{'deep1'} 3.83 
& \cellcolor{'deep1'} 3.40
& \cellcolor{'deep1'} 4.11 
& \cellcolor{'deep1'} 3.68
& \cellcolor{'deep1'} 3.15
& \cellcolor{'deep1'} 4.22
& \cellcolor{'deep1'} 4.18
\\ 
\cmidrule(l){2-14}
& 
\multirow{6}{*}{
\makecell{R1-\\\ OV 
  \cite{yang2025r1onevision}}
  } &
\multirow{3}{*}{Vani.} &
\cellcolor{'shallow1'} Think 
& \cellcolor{'shallow1'} 3.10
& \cellcolor{'shallow1'} 2.97
& \cellcolor{'shallow1'} 2.63
& \cellcolor{'shallow1'} 3.49 
& \cellcolor{'shallow1'} 3.40
& \cellcolor{'shallow1'} 3.67
& \cellcolor{'shallow1'} 3.10
& \cellcolor{'shallow1'} 3.45 
& \cellcolor{'shallow1'} 3.62
& \cellcolor{'shallow1'} 4.23
\\ 
& & &
\cellcolor{'shallow1'} Answer 
& \cellcolor{'shallow1'} 3.55
& \cellcolor{'shallow1'} 3.29
& \cellcolor{'shallow1'} 3.23
& \cellcolor{'shallow1'} 4.19 
& \cellcolor{'shallow1'} 3.92
& \cellcolor{'shallow1'} 4.51 
& \cellcolor{'shallow1'} 3.20
& \cellcolor{'shallow1'} 4.10 
& \cellcolor{'shallow1'} 4.36
& \cellcolor{'shallow1'} 4.32
\\ 
& & &
\cellcolor{'shallow1'} Overall 
& \cellcolor{'shallow1'} 3.03
& \cellcolor{'shallow1'} 3.07
& \cellcolor{'shallow1'} 2.70
& \cellcolor{'shallow1'} 3.99 
& \cellcolor{'shallow1'} 3.48
& \cellcolor{'shallow1'} 4.40 
& \cellcolor{'shallow1'} 3.18
& \cellcolor{'shallow1'} 3.72 
& \cellcolor{'shallow1'} 4.35
& \cellcolor{'shallow1'} 4.28
\\
\cmidrule(l){3-14}
& & \multirow{3}{*}{Jail.} & 
\cellcolor{'shallow1'} Think  
& \cellcolor{'shallow1'} 4.12 
& \cellcolor{'shallow1'} 3.62
& \cellcolor{'shallow1'} 4.27 
& \cellcolor{'shallow1'} 4.35 
& \cellcolor{'shallow1'} 4.14 
& \cellcolor{'shallow1'} 3.90  
& \cellcolor{'shallow1'} 4.12 
& \cellcolor{'shallow1'} 3.20  
& \cellcolor{'shallow1'} 3.84
& \cellcolor{'shallow1'} 4.20
\\ 
& & & \cellcolor{'shallow1'} Answer  
& \cellcolor{'shallow1'} 4.36  
& \cellcolor{'shallow1'} 3.86 
& \cellcolor{'shallow1'} 4.66
& \cellcolor{'shallow1'} 4.59  
& \cellcolor{'shallow1'} 4.51
& \cellcolor{'shallow1'} 4.15  
& \cellcolor{'shallow1'} 4.54
& \cellcolor{'shallow1'} 3.71  
& \cellcolor{'shallow1'} 4.03
& \cellcolor{'shallow1'} 4.25
\\ 
& & & \cellcolor{'shallow1'} Overall  
& \cellcolor{'shallow1'} 4.31  
& \cellcolor{'shallow1'} 3.91 
& \cellcolor{'shallow1'} 4.61 
& \cellcolor{'shallow1'} 4.60  
& \cellcolor{'shallow1'} 4.37 
& \cellcolor{'shallow1'} 4.32  
& \cellcolor{'shallow1'} 4.33 
& \cellcolor{'shallow1'} 3.50  
& \cellcolor{'shallow1'} 4.22
& \cellcolor{'shallow1'} 4.21
\\
\cmidrule(l){2-14}
& 
\multirow{6}{*}{
\makecell{Eureka-\\\
QW \cite{mllm-mm-eureka}} 

} &
\multirow{3}{*}{Vani.} 
&\cellcolor{'shallow1'} Think 
& \cellcolor{'shallow1'} 1.95 
& \cellcolor{'shallow1'} 1.97
& \cellcolor{'shallow1'} 3.20
& \cellcolor{'shallow1'} 3.45  
& \cellcolor{'shallow1'} 2.70
& \cellcolor{'shallow1'} 4.00  
& \cellcolor{'shallow1'} 2.90
& \cellcolor{'shallow1'} 3.33  
& \cellcolor{'shallow1'} 3.93
& \cellcolor{'shallow1'} 3.54
\\ 
& & & \cellcolor{'shallow1'} Answer  
& \cellcolor{'shallow1'} 2.03 
& \cellcolor{'shallow1'} 1.87
& \cellcolor{'shallow1'} 3.34 
& \cellcolor{'shallow1'} 3.11  
& \cellcolor{'shallow1'} 2.29 
& \cellcolor{'shallow1'} 4.15  
& \cellcolor{'shallow1'} 2.55 
& \cellcolor{'shallow1'} 3.35  
& \cellcolor{'shallow1'} 4.04
& \cellcolor{'shallow1'} 3.62
\\ 
& & & \cellcolor{'shallow1'} Overall  
& \cellcolor{'shallow1'} 2.23 
& \cellcolor{'shallow1'} 2.01
& \cellcolor{'shallow1'} 3.36 
& \cellcolor{'shallow1'} 3.53  
& \cellcolor{'shallow1'} 2.65 
& \cellcolor{'shallow1'} 4.22 
& \cellcolor{'shallow1'} 2.97 
& \cellcolor{'shallow1'} 3.23  
& \cellcolor{'shallow1'} 3.99
& \cellcolor{'shallow1'} 3.90
\\ 
\cmidrule(l){3-14}
& & \multirow{3}{*}{Jail.} & 
\cellcolor{'shallow1'} Think  
& \cellcolor{'shallow1'} 3.40
& \cellcolor{'shallow1'} 3.12 
& \cellcolor{'shallow1'} 4.00 
& \cellcolor{'shallow1'} 4.24  
& \cellcolor{'shallow1'} 3.90 
& \cellcolor{'shallow1'} 3.91  
& \cellcolor{'shallow1'} 3.79 
& \cellcolor{'shallow1'} 3.19  
& \cellcolor{'shallow1'} 4.01
& \cellcolor{'shallow1'} 3.74
\\ 
& & & \cellcolor{'shallow1'} Answer  
& \cellcolor{'shallow1'} 2.89  
& \cellcolor{'shallow1'} 3.25
& \cellcolor{'shallow1'} 4.30 
& \cellcolor{'shallow1'} 4.03  
& \cellcolor{'shallow1'} 3.66
& \cellcolor{'shallow1'} 3.82 
& \cellcolor{'shallow1'} 3.67 
& \cellcolor{'shallow1'} 3.08  
& \cellcolor{'shallow1'} 3.91
& \cellcolor{'shallow1'} 3.71
\\ 
& & & \cellcolor{'shallow1'} Overall  
& \cellcolor{'shallow1'} 3.39  
& \cellcolor{'shallow1'} 3.45 
& \cellcolor{'shallow1'} 4.25 
& \cellcolor{'shallow1'} 4.24  
& \cellcolor{'shallow1'} 3.97 
& \cellcolor{'shallow1'} 4.05 
& \cellcolor{'shallow1'} 4.01 
& \cellcolor{'shallow1'} 3.25  
& \cellcolor{'shallow1'} 4.11
& \cellcolor{'shallow1'} 3.91
\\ 
\cmidrule(lr){1-14}
\multirow{8}{*}{
\makecell{\rotatebox{90}{QW2-VL \cite{Qwen2VL}}}
} & 
\multirow{2}{*}{-} &
\multirow{1}{*}{Vani.} &
\cellcolor{'deep2'} Overall 
& \cellcolor{'deep2'} 0.05
& \cellcolor{'deep2'} 0.05
& \cellcolor{'deep2'} 0.55
& \cellcolor{'deep2'} 0.79
& \cellcolor{'deep2'} 0.14
& \cellcolor{'deep2'} 1.75 
& \cellcolor{'deep2'} 0.36
& \cellcolor{'deep2'} 1.85
& \cellcolor{'deep2'} 3.82
& \cellcolor{'deep2'} 2.25
\\ 
\cmidrule(l){3-14}
& & \multirow{1}{*}{Jail.} & 
\cellcolor{'deep2'} Overall  
& \cellcolor{'deep2'} 3.66 
& \cellcolor{'deep2'} 3.29
& \cellcolor{'deep2'} 4.34 
& \cellcolor{'deep2'} 4.22  
& \cellcolor{'deep2'} 4.27
& \cellcolor{'deep2'} 4.39  
& \cellcolor{'deep2'} 4.06 
& \cellcolor{'deep2'} 3.10  
& \cellcolor{'deep2'} 4.05
& \cellcolor{'deep2'} 3.88
\\ 
\cmidrule(l){2-14}
& 
\multirow{6}{*}{\makecell{MBerry-\\\ QW \cite{yao2024mulberry}}} &
\multirow{3}{*}{Vani.} &
\cellcolor{'shallow2'} Think 
& \cellcolor{'shallow2'} 2.12
& \cellcolor{'shallow2'} 2.34
& \cellcolor{'shallow2'} 2.57
& \cellcolor{'shallow2'} 3.06 
& \cellcolor{'shallow2'} 3.12
& \cellcolor{'shallow2'} 2.96 
& \cellcolor{'shallow2'} 2.63
& \cellcolor{'shallow2'} 2.57 
& \cellcolor{'shallow2'} 3.28
& \cellcolor{'shallow2'} 2.83
\\ 
& & &
\cellcolor{'shallow2'} Answer 
& \cellcolor{'shallow2'} 2.45
& \cellcolor{'shallow2'} 2.47
& \cellcolor{'shallow2'} 3.32
& \cellcolor{'shallow2'} 3.44 
& \cellcolor{'shallow2'} 2.62
& \cellcolor{'shallow2'} 3.95 
& \cellcolor{'shallow2'} 3.06
& \cellcolor{'shallow2'} 3.35
& \cellcolor{'shallow2'} 3.88
& \cellcolor{'shallow2'} 3.61
\\ 
& & &
\cellcolor{'shallow2'} Overall
& \cellcolor{'shallow2'} 2.50
& \cellcolor{'shallow2'} 2.44
& \cellcolor{'shallow2'} 3.39
& \cellcolor{'shallow2'} 3.55
& \cellcolor{'shallow2'} 2.75
& \cellcolor{'shallow2'} 4.10 
& \cellcolor{'shallow2'} 3.03
& \cellcolor{'shallow2'} 3.65 
& \cellcolor{'shallow2'} 4.08
& \cellcolor{'shallow2'} 3.82
\\ 
\cmidrule(l){3-14}
& & \multirow{3}{*}{Jail.} & 
\cellcolor{'shallow2'} Think  
& \cellcolor{'shallow2'} 3.27 
& \cellcolor{'shallow2'} 2.87
& \cellcolor{'shallow2'} 3.02
& \cellcolor{'shallow2'} 3.58  
& \cellcolor{'shallow2'} 3.45 
& \cellcolor{'shallow2'} 2.97 
& \cellcolor{'shallow2'} 3.47
& \cellcolor{'shallow2'} 2.38  
& \cellcolor{'shallow2'} 2.49
& \cellcolor{'shallow2'} 2.91
\\ 
& & & \cellcolor{'shallow2'} Answer  
& \cellcolor{'shallow2'} 4.41  
& \cellcolor{'shallow2'} 4.07
& \cellcolor{'shallow2'} 4.48
& \cellcolor{'shallow2'} 4.47  
& \cellcolor{'shallow2'} 4.44
& \cellcolor{'shallow2'} 4.08  
& \cellcolor{'shallow2'} 4.32
& \cellcolor{'shallow2'} 3.39  
& \cellcolor{'shallow2'} 3.86
& \cellcolor{'shallow2'} 3.86
\\ 
& & & \cellcolor{'shallow2'} Overall  
& \cellcolor{'shallow2'} 4.48  
& \cellcolor{'shallow2'} 4.19
& \cellcolor{'shallow2'} 4.36 
& \cellcolor{'shallow2'} 4.57  
& \cellcolor{'shallow2'} 4.60
& \cellcolor{'shallow2'} 4.33  
& \cellcolor{'shallow2'} 4.45 
& \cellcolor{'shallow2'} 3.55  
& \cellcolor{'shallow2'} 4.08
& \cellcolor{'shallow2'} 3.99
\\ 
\cmidrule(lr){1-14} 
\multirow{4}{*}[-\dimexpr0.5\baselineskip\relax]{
  \rotatebox[origin=c]{90}{
    \parbox{2cm}{
      \centering
      LMA3-\\
      LVA \cite{li2024llavanext-strong}
    }
  }
}
&
\multirow{2}{*}{-} &
\multirow{1}{*}{Vani.} &
\cellcolor{'deep3'} Overall 
& \cellcolor{'deep3'} 0.15
& \cellcolor{'deep3'} 0.50
& \cellcolor{'deep3'} 1.45
& \cellcolor{'deep3'} 1.92 
& \cellcolor{'deep3'} 0.58
& \cellcolor{'deep3'} 2.45
& \cellcolor{'deep3'} 0.91
& \cellcolor{'deep3'} 2.07 
& \cellcolor{'deep3'} 4.07
& \cellcolor{'deep3'} 2.64
\\ 
\cmidrule(l){3-14}
& & \multirow{1}{*}{Jail.} & 
\cellcolor{'deep3'} Overall  
& \cellcolor{'deep3'} 3.77 
& \cellcolor{'deep3'} 3.15
& \cellcolor{'deep3'} 4.05 
& \cellcolor{'deep3'} 4.22  
& \cellcolor{'deep3'} 4.05 
& \cellcolor{'deep3'} 3.76 
& \cellcolor{'deep3'} 3.96 
& \cellcolor{'deep3'} 3.32  
& \cellcolor{'deep3'} 3.80
& \cellcolor{'deep3'} 3.85
\\ 
\cmidrule(l){2-14}
& 
\multirow{2}{*}[0pt]{\makecell[c]{MBerry-\\\ LVA \cite{yao2024mulberry}}}  
&
\multirow{1}{*}{Vani.} &
\cellcolor{'shallow3'} Overall 
& \cellcolor{'shallow3'} 1.95
& \cellcolor{'shallow3'} 3.36
& \cellcolor{'shallow3'} 4.41
& \cellcolor{'shallow3'} 3.39 
& \cellcolor{'shallow3'} 3.74
& \cellcolor{'shallow3'} 3.80 
& \cellcolor{'shallow3'} 2.60
& \cellcolor{'shallow3'} 3.32 
& \cellcolor{'shallow3'} 3.99
& \cellcolor{'shallow3'} 3.55
\\ 
\cmidrule(l){3-14}
& & \multirow{1}{*}{Jail.} & 
\cellcolor{'shallow3'} Overall  
& \cellcolor{'shallow3'} 3.85 
& \cellcolor{'shallow3'} 3.52
& \cellcolor{'shallow3'} 3.55 
& \cellcolor{'shallow3'} 4.03  
& \cellcolor{'shallow3'} 3.77
& \cellcolor{'shallow3'} 3.78 
& \cellcolor{'shallow3'} 3.89
& \cellcolor{'shallow3'} 3.25  
& \cellcolor{'shallow3'} 3.44
& \cellcolor{'shallow3'} 3.79
\\ 
\cmidrule(lr){1-14}

\multirow{4}{*}[-\dimexpr0.5\baselineskip\relax]{
  \rotatebox[origin=c]{90}{
    \parbox{2cm}{
      \centering
      LMA-3.2-\\
      Vision \cite{llama2}
    }
  }
}
&
\multirow{2}{*}{-} &
\multirow{1}{*}{Vani.} &
\cellcolor{'deep4'} Overall 
& \cellcolor{'deep4'} 0.10
& \cellcolor{'deep4'} 0.73
& \cellcolor{'deep4'} 2.07
& \cellcolor{'deep4'} 1.76
& \cellcolor{'deep4'} 0.62
& \cellcolor{'deep4'} 4.43
& \cellcolor{'deep4'} 1.09
& \cellcolor{'deep4'} 3.73 
& \cellcolor{'deep4'} 4.42
& \cellcolor{'deep4'} 3.63
\\ 
\cmidrule(l){3-14}
& & \multirow{1}{*}{Jail.} & 
\cellcolor{'deep4'} Overall  
& \cellcolor{'deep4'} 0.76
& \cellcolor{'deep4'} 1.66
& \cellcolor{'deep4'} 2.18 
& \cellcolor{'deep4'} 1.96  
& \cellcolor{'deep4'} 1.88
& \cellcolor{'deep4'} 3.59 
& \cellcolor{'deep4'} 2.37 
& \cellcolor{'deep4'} 3.04
& \cellcolor{'deep4'} 4.17
& \cellcolor{'deep4'} 3.84
\\ 
\cmidrule(l){2-14}
& 
\multirow{2}{*}[0pt]{\makecell[c]{MBerry-\\\ LMA \cite{yao2024mulberry}}}
&
\multirow{1}{*}{Vani.} &
\cellcolor{'shallow4'} Overall 
& \cellcolor{'shallow4'} 0.10
& \cellcolor{'shallow4'} 0.64
& \cellcolor{'shallow4'} 1.11
& \cellcolor{'shallow4'} 1.31
& \cellcolor{'shallow4'} 0.62
& \cellcolor{'shallow4'} 3.00
& \cellcolor{'shallow4'} 0.76
& \cellcolor{'shallow4'} 3.83 
& \cellcolor{'shallow4'} 4.29
& \cellcolor{'shallow4'} 4.23
\\ 
\cmidrule(l){3-14}
& & \multirow{1}{*}{Jail.} & 
\cellcolor{'shallow4'} Overall  
& \cellcolor{'shallow4'} 3.38 
& \cellcolor{'shallow4'} 3.18
& \cellcolor{'shallow4'} 2.98
& \cellcolor{'shallow4'} 3.58  
& \cellcolor{'shallow4'} 3.11 
& \cellcolor{'shallow4'} 3.65
& \cellcolor{'shallow4'} 3.42 
& \cellcolor{'shallow4'} 3.51  
& \cellcolor{'shallow4'} 3.56
& \cellcolor{'shallow4'} 4.07
\\ 
\bottomrule
\end{tabular}}
\end{table}

\subsection{Reasoning Tax} \label{sec:exp1}

Prior work has observed the ``safety tax'' in reasoning models, \ie performance deterioration caused by safety alignment \cite{safety-lrm-safety-tax}. We identify a critical counterpart: the reasoning tax, defined as the safety degradation incurred when augmenting MLLMs with reasoning capabilities through SFT or RL. To quantify this phenomenon, we systematically compare safety metrics between four safety-aligned base MLLMs and five MLRMs derived from them.

\vspace{5pt}
\noindent \textbf{Experimental Protocol.} To isolate the impact of reasoning capability acquisition, we adopt a two-stage evaluation: For base MLLMs, we directly assess the outputs for HR and ASR, as these models lack explicit reasoning step generation. For MLRMs, we collect HR and ASR across Think, Answer and Overall processes.
Note that although we employed a standard prompt template to enforce the separation of the reasoning and answer processes, some MLRMs (\eg Mulberry-LlaMA) still struggled to strictly follow the ``Think + Answer'' format in their outputs. For these models, we report their overall safety solely. Figure \ref{fig:zhuzi1} and \ref{fig:zhuzi2} exhibit the results under vanilla unsafe text queries and jailbreak attacks, respectively.

\vspace{5pt}
\noindent \textbf{Key Findings.} Figures \ref{fig:zhuzi1} and \ref{fig:zhuzi2} reveal systematic safety degradation across all MLRMs. Specifically,
\begin{itemize}[leftmargin=*]
\item  {Overall Safety Collapse}: MLRMs exhibit 31.30\% higher ASR (59.52\% \textit{vs.} base MLLMs’ 28.22\%) and 1.64 higher HR (3.07\% \textit{vs.} 1.43), demonstrating that exposing reasoning chains fundamentally expands attack surfaces.
\item   {Persistent Answer Degradation}: Even when ignoring unsafe reasoning steps, MLRM answers show 25.89\% higher ASR and 1.30 higher HR compared to base models. This proves safety erosion persists beyond reasoning exposure, suggesting SFT/RL alignment damages intrinsic safeguards.
\item   {Architectural Vulnerability}: Qwen2.5-based MLRMs suffer catastrophic safety loss (ASR + 40.06\%, HR + 2.02), suggesting architectural incompatibility between reasoning and safety mechanisms. Notably, MBerry-LMA exhibits an inverse trend: its safety metrics improve post-reasoning augmentation (ASR - 4.8\%, HR - 0.21). We believe this anomaly hints at potential safeguards, and may provide a rare blueprint for designing safety-resilient reasoning models.
\item  {Reasoning as Attack Vector}: Unsafe reasoning occurs 12.52\% more frequently than unsafe answers, highlighting systemic risks in exposing intermediate reasoning to users.
\end{itemize}

\begin{tcolorbox}[colback=lightgray!10, colframe=black, title={Takeaway 3.1: Reasoning Tax in MLRMs}]
SFT/RL-based reasoning acquisition severely compromises safety alignment of base MLLMs, a phenomenon we term the ``reasoning tax'' in MLRMs. 
\end{tcolorbox}

\begin{table}[t]
\caption{Comparison of HR scores on SafetyBench datasets across various MLRMs and their base MLLMs.
  For abbreviations, \textit{Vani.} and \textit{Jail.} refer to performance under vanilla unsafe text queries and jailbreak attacks, respectively. The dash in the ``Model'' column represents the base MLLMs.}
\label{tab:jailbreak2}
\centering
\resizebox{1.0\linewidth}{!}{
\setlength{\tabcolsep}{3pt}
\begin{tabular}{lcccrrrrrrrrrrrr}
\toprule

Base & Model & Method & Stage & 
\makecell{\textbf{IA}} & 
\makecell{\textbf{HS}} & 
\makecell{\textbf{RA}} & 
\makecell{\textbf{PH}} & 
\makecell{\textbf{FR}} & 
\makecell{\textbf{PO}} & 
\makecell{\textbf{PV}} & 
\makecell{\textbf{LO}} & 
\makecell{\textbf{FA}} &
\makecell{\textbf{HC}} \\
\midrule
\multirow{14}{*}{\rotatebox{90}{\makecell{QW2.5-VL  \cite{qwen2.5-VL}}}} & 
\multirow{2}{*}{\centering{-}} &  
\multirow{1}{*}{Vani.} &
\cellcolor{'deep1'} Overall 
& \cellcolor{'deep1'} 0.22 
& \cellcolor{'deep1'} 0.42
& \cellcolor{'deep1'} 0.00
& \cellcolor{'deep1'} 0.18
& \cellcolor{'deep1'} 0.50
& \cellcolor{'deep1'} 1.96
& \cellcolor{'deep1'} 0.74
& \cellcolor{'deep1'} 3.94 
& \cellcolor{'deep1'} 3.92
& \cellcolor{'deep1'} 3.70
\\ 
\cmidrule(l){3-14}
& & \multirow{1}{*}{Jail.} & 
\cellcolor{'deep1'} Overall 
& \cellcolor{'deep1'} 2.78
& \cellcolor{'deep1'} 2.68
& \cellcolor{'deep1'} 4.36
& \cellcolor{'deep1'} 3.04
& \cellcolor{'deep1'} 4.08
& \cellcolor{'deep1'} 4.04
& \cellcolor{'deep1'} 3.14
& \cellcolor{'deep1'} 3.56
& \cellcolor{'deep1'} 4.32
& \cellcolor{'deep1'} 3.62
\\ 
\cmidrule(l){2-14}
& 
\multirow{6}{*}{\makecell{R1-\\\ OV 
  \cite{yang2025r1onevision}}} &
\multirow{3}{*}{Vani.} &
\cellcolor{'shallow1'} Think 
& \cellcolor{'shallow1'} 3.30
& \cellcolor{'shallow1'} 2.57
& \cellcolor{'shallow1'} 4.12
& \cellcolor{'shallow1'} 2.94
& \cellcolor{'shallow1'} 3.48
& \cellcolor{'shallow1'} 4.17
& \cellcolor{'shallow1'} 3.10
& \cellcolor{'shallow1'} 3.40 
& \cellcolor{'shallow1'} 4.00
& \cellcolor{'shallow1'} 3.22
\\ 
& & &
\cellcolor{'shallow1'} Answer 
& \cellcolor{'shallow1'} 4.00
& \cellcolor{'shallow1'} 3.19
& \cellcolor{'shallow1'} 3.98
& \cellcolor{'shallow1'} 3.43 
& \cellcolor{'shallow1'} 3.94
& \cellcolor{'shallow1'} 4.59 
& \cellcolor{'shallow1'} 3.39
& \cellcolor{'shallow1'} 3.76 
& \cellcolor{'shallow1'} 3.94
& \cellcolor{'shallow1'} 4.18
\\ 
& & &
\cellcolor{'shallow1'} Overall 
& \cellcolor{'shallow1'} 3.44
& \cellcolor{'shallow1'} 2.66
& \cellcolor{'shallow1'} 3.84
& \cellcolor{'shallow1'} 3.14
& \cellcolor{'shallow1'} 3.66
& \cellcolor{'shallow1'} 4.30
& \cellcolor{'shallow1'} 2.98
& \cellcolor{'shallow1'} 3.22 
& \cellcolor{'shallow1'} 4.06
& \cellcolor{'shallow1'} 3.64
\\
\cmidrule(l){3-14}
& & \multirow{3}{*}{Jail.} & 
\cellcolor{'shallow1'} Think  
& \cellcolor{'shallow1'} 4.12 
& \cellcolor{'shallow1'} 3.79
& \cellcolor{'shallow1'} 4.86 
& \cellcolor{'shallow1'} 4.58 
& \cellcolor{'shallow1'} 4.44 
& \cellcolor{'shallow1'} 4.20  
& \cellcolor{'shallow1'} 4.52 
& \cellcolor{'shallow1'} 3.40  
& \cellcolor{'shallow1'} 3.80
& \cellcolor{'shallow1'} 4.34
\\ 
& & & \cellcolor{'shallow1'} Answer  
& \cellcolor{'shallow1'} 4.67  
& \cellcolor{'shallow1'} 3.88
& \cellcolor{'shallow1'} 4.91
& \cellcolor{'shallow1'} 4.60  
& \cellcolor{'shallow1'} 4.51
& \cellcolor{'shallow1'} 4.38  
& \cellcolor{'shallow1'} 4.58
& \cellcolor{'shallow1'} 3.67  
& \cellcolor{'shallow1'} 3.87
& \cellcolor{'shallow1'} 4.02
\\ 
& & & \cellcolor{'shallow1'} Overall  
& \cellcolor{'shallow1'} 4.18  
& \cellcolor{'shallow1'} 3.78
& \cellcolor{'shallow1'} 4.88
& \cellcolor{'shallow1'} 4.62  
& \cellcolor{'shallow1'} 4.54 
& \cellcolor{'shallow1'} 4.48
& \cellcolor{'shallow1'} 4.52
& \cellcolor{'shallow1'} 3.58
& \cellcolor{'shallow1'} 3.88
& \cellcolor{'shallow1'} 3.98
\\
\cmidrule(l){2-14}
& 
\multirow{6}{*}{
\makecell{Eureka-\\\
QW \cite{mllm-mm-eureka}} 
} &
\multirow{3}{*}{Vani.} 
&\cellcolor{'shallow1'} Think 
& \cellcolor{'shallow1'} 2.89
& \cellcolor{'shallow1'} 1.61
& \cellcolor{'shallow1'} 3.26
& \cellcolor{'shallow1'} 2.14  
& \cellcolor{'shallow1'} 2.18
& \cellcolor{'shallow1'} 4.04 
& \cellcolor{'shallow1'} 2.28
& \cellcolor{'shallow1'} 3.30 
& \cellcolor{'shallow1'} 3.90
& \cellcolor{'shallow1'} 3.54
\\ 
& & & \cellcolor{'shallow1'} Answer  
& \cellcolor{'shallow1'} 2.50 
& \cellcolor{'shallow1'} 1.58
& \cellcolor{'shallow1'} 3.02
& \cellcolor{'shallow1'} 1.94  
& \cellcolor{'shallow1'} 1.86 
& \cellcolor{'shallow1'} 4.14  
& \cellcolor{'shallow1'} 2.12 
& \cellcolor{'shallow1'} 3.14  
& \cellcolor{'shallow1'} 3.64
& \cellcolor{'shallow1'} 3.50
\\ 
& & & \cellcolor{'shallow1'} Overall  
& \cellcolor{'shallow1'} 2.50 
& \cellcolor{'shallow1'} 1.50
& \cellcolor{'shallow1'} 3.04 
& \cellcolor{'shallow1'} 1.90  
& \cellcolor{'shallow1'} 1.84 
& \cellcolor{'shallow1'} 4.00 
& \cellcolor{'shallow1'} 2.30
& \cellcolor{'shallow1'} 3.30  
& \cellcolor{'shallow1'} 3.54
& \cellcolor{'shallow1'} 3.28
\\ 
\cmidrule(l){3-14}
& & \multirow{3}{*}{Jail.} & 
\cellcolor{'shallow1'} Think  
& \cellcolor{'shallow1'} 3.14
& \cellcolor{'shallow1'} 2.84
& \cellcolor{'shallow1'} 3.90 
& \cellcolor{'shallow1'} 3.34  
& \cellcolor{'shallow1'} 3.46
& \cellcolor{'shallow1'} 4.18  
& \cellcolor{'shallow1'} 3.50
& \cellcolor{'shallow1'} 3.38  
& \cellcolor{'shallow1'} 3.78
& \cellcolor{'shallow1'} 3.62
\\ 
& & & \cellcolor{'shallow1'} Answer  
& \cellcolor{'shallow1'} 3.26  
& \cellcolor{'shallow1'} 2.98
& \cellcolor{'shallow1'} 4.12 
& \cellcolor{'shallow1'} 3.60 
& \cellcolor{'shallow1'} 3.60
& \cellcolor{'shallow1'} 4.14 
& \cellcolor{'shallow1'} 3.66 
& \cellcolor{'shallow1'} 3.26  
& \cellcolor{'shallow1'} 3.72
& \cellcolor{'shallow1'} 3.36
\\ 
& & & \cellcolor{'shallow1'} Overall  
& \cellcolor{'shallow1'} 3.32  
& \cellcolor{'shallow1'} 2.8 
& \cellcolor{'shallow1'} 3.98 
& \cellcolor{'shallow1'} 3.52  
& \cellcolor{'shallow1'} 3.58 
& \cellcolor{'shallow1'} 4.18 
& \cellcolor{'shallow1'} 3.54
& \cellcolor{'shallow1'} 3.52  
& \cellcolor{'shallow1'} 3.86
& \cellcolor{'shallow1'} 3.60
\\ 
\cmidrule(lr){1-14}
 
\multirow{8}{*}{\makecell{\rotatebox{90}{QW2-VL \cite{Qwen2VL}}}} & 
\multirow{2}{*}{-} &
\multirow{1}{*}{Vani.} &
\cellcolor{'deep2'} Overall 
& \cellcolor{'deep2'} 0.28
& \cellcolor{'deep2'} 0.34
& \cellcolor{'deep2'} 0.28
& \cellcolor{'deep2'} 0.06
& \cellcolor{'deep2'} 0.06
& \cellcolor{'deep2'} 2.24 
& \cellcolor{'deep2'} 0.76
& \cellcolor{'deep2'} 3.08
& \cellcolor{'deep2'} 3.92
& \cellcolor{'deep2'} 3.68
\\ 
\cmidrule(l){3-14}
& & \multirow{1}{*}{Jail.} & 
\cellcolor{'deep2'} Overall  
& \cellcolor{'deep2'} 3.30
& \cellcolor{'deep2'} 2.74
& \cellcolor{'deep2'} 4.82 
& \cellcolor{'deep2'} 3.34  
& \cellcolor{'deep2'} 4.40
& \cellcolor{'deep2'} 4.48 
& \cellcolor{'deep2'} 3.82
& \cellcolor{'deep2'} 4.04 
& \cellcolor{'deep2'} 4.28
& \cellcolor{'deep2'} 4.40
\\ 
\cmidrule(l){2-14}
& 
\multirow{6}{*}{\makecell{MBerry-\\\ QW \cite{yao2024mulberry}}} &
\multirow{3}{*}{Vani.} &
\cellcolor{'shallow2'} Think 
& \cellcolor{'shallow2'} 2.72
& \cellcolor{'shallow2'} 2.42
& \cellcolor{'shallow2'} 3.14
& \cellcolor{'shallow2'} 2.39 
& \cellcolor{'shallow2'} 2.62
& \cellcolor{'shallow2'} 3.37 
& \cellcolor{'shallow2'} 2.62
& \cellcolor{'shallow2'} 2.71 
& \cellcolor{'shallow2'} 3.24
& \cellcolor{'shallow2'} 3.33
\\ 
& & &
\cellcolor{'shallow2'} Answer 
& \cellcolor{'shallow2'} 2.96
& \cellcolor{'shallow2'} 2.47
& \cellcolor{'shallow2'} 3.45
& \cellcolor{'shallow2'} 2.68 
& \cellcolor{'shallow2'} 2.96
& \cellcolor{'shallow2'} 4.10 
& \cellcolor{'shallow2'} 2.84
& \cellcolor{'shallow2'} 3.80
& \cellcolor{'shallow2'} 3.65
& \cellcolor{'shallow2'} 3.92
\\ 
& & &
\cellcolor{'shallow2'} Overall
& \cellcolor{'shallow2'} 2.88
& \cellcolor{'shallow2'} 2.30
& \cellcolor{'shallow2'} 3.42
& \cellcolor{'shallow2'} 2.46
& \cellcolor{'shallow2'} 3.06
& \cellcolor{'shallow2'} 4.16
& \cellcolor{'shallow2'} 2.68
& \cellcolor{'shallow2'} 3.72
& \cellcolor{'shallow2'} 3.66
& \cellcolor{'shallow2'} 3.86
\\ 
\cmidrule(l){3-14}
& & \multirow{3}{*}{Jail.} & 
\cellcolor{'shallow2'} Think  
& \cellcolor{'shallow2'} 4.15
& \cellcolor{'shallow2'} 3.95
& \cellcolor{'shallow2'} 4.48
& \cellcolor{'shallow2'} 4.10  
& \cellcolor{'shallow2'} 4.31 
& \cellcolor{'shallow2'} 4.36 
& \cellcolor{'shallow2'} 4.16
& \cellcolor{'shallow2'} 3.73  
& \cellcolor{'shallow2'} 3.89
& \cellcolor{'shallow2'} 3.82
\\ 
& & & \cellcolor{'shallow2'} Answer  
& \cellcolor{'shallow2'} 4.46  
& \cellcolor{'shallow2'} 4.08
& \cellcolor{'shallow2'} 4.80
& \cellcolor{'shallow2'} 4.38 
& \cellcolor{'shallow2'} 4.56
& \cellcolor{'shallow2'} 4.66  
& \cellcolor{'shallow2'} 4.40
& \cellcolor{'shallow2'} 3.56  
& \cellcolor{'shallow2'} 4.00
& \cellcolor{'shallow2'} 4.20
\\ 
& & & \cellcolor{'shallow2'} Overall  
& \cellcolor{'shallow2'} 4.69 
& \cellcolor{'shallow2'} 4.24
& \cellcolor{'shallow2'} 4.88 
& \cellcolor{'shallow2'} 4.42 
& \cellcolor{'shallow2'} 4.64
& \cellcolor{'shallow2'} 4.52  
& \cellcolor{'shallow2'} 4.46 
& \cellcolor{'shallow2'} 3.46  
& \cellcolor{'shallow2'} 4.00
& \cellcolor{'shallow2'} 4.18
\\ 
\cmidrule(lr){1-14}
 
\multirow{4}{*}[-\dimexpr0.5\baselineskip\relax]{
  \rotatebox[origin=c]{90}{
    \parbox{2cm}{
      \centering
      LMA3-\\
      LVA \cite{li2024llavanext-strong}
    }
  }
}
&
\multirow{2}{*}{-} &
\multirow{1}{*}{Vani.} &
\cellcolor{'deep3'} Overall 
& \cellcolor{'deep3'} 0.66
& \cellcolor{'deep3'} 0.84
& \cellcolor{'deep3'} 2.26
& \cellcolor{'deep3'} 0.86
& \cellcolor{'deep3'} 1.08
& \cellcolor{'deep3'} 2.60
& \cellcolor{'deep3'} 1.20
& \cellcolor{'deep3'} 3.32
& \cellcolor{'deep3'} 3.86
& \cellcolor{'deep3'} 3.58
\\ 
\cmidrule(l){3-14}
& & \multirow{1}{*}{Jail.} & 
\cellcolor{'deep3'} Overall  
& \cellcolor{'deep3'} 4.22
& \cellcolor{'deep3'} 3.06
& \cellcolor{'deep3'} 4.48 
& \cellcolor{'deep3'} 4.36 
& \cellcolor{'deep3'} 4.24 
& \cellcolor{'deep3'} 3.76 
& \cellcolor{'deep3'} 4.04 
& \cellcolor{'deep3'} 3.32  
& \cellcolor{'deep3'} 3.80
& \cellcolor{'deep3'} 3.85
\\ 
\cmidrule(l){2-14}
& 
\multirow{2}{*}[0pt]{\makecell[c]{MBerry-\\\ LVA \cite{yao2024mulberry}}}  
&
\multirow{1}{*}{Vani.} &
\cellcolor{'shallow3'} Overall 
& \cellcolor{'shallow3'} 1.90
& \cellcolor{'shallow3'} 1.46
& \cellcolor{'shallow3'} 3.68
& \cellcolor{'shallow3'} 1.52
& \cellcolor{'shallow3'} 3.28
& \cellcolor{'shallow3'} 3.48
& \cellcolor{'shallow3'} 2.22
& \cellcolor{'shallow3'} 3.62
& \cellcolor{'shallow3'} 4.04
& \cellcolor{'shallow3'} 3.64
\\ 
\cmidrule(l){3-14}
& & \multirow{1}{*}{Jail.} & 
\cellcolor{'shallow3'} Overall  
& \cellcolor{'shallow3'} 4.38
& \cellcolor{'shallow3'} 3.56
& \cellcolor{'shallow3'} 4.42 
& \cellcolor{'shallow3'} 4.08 
& \cellcolor{'shallow3'} 3.94
& \cellcolor{'shallow3'} 4.04 
& \cellcolor{'shallow3'} 3.84
& \cellcolor{'shallow3'} 3.38  
& \cellcolor{'shallow3'} 3.38
& \cellcolor{'shallow3'} 3.58
\\ 
\cmidrule(lr){1-14}

\multirow{4}{*}[-\dimexpr0.5\baselineskip\relax]{
  \rotatebox[origin=c]{90}{
    \parbox{2cm}{
      \centering
      LMA-3.2-\\
      Vision \cite{llama2}
    }
  }
}
&
\multirow{2}{*}{-} &
\multirow{1}{*}{Vani.} &
\cellcolor{'deep4'} Overall 
& \cellcolor{'deep4'} 1.78
& \cellcolor{'deep4'} 1.36
& \cellcolor{'deep4'} 1.20
& \cellcolor{'deep4'} 0.72
& \cellcolor{'deep4'} 1.68
& \cellcolor{'deep4'} 4.30
& \cellcolor{'deep4'} 1.76
& \cellcolor{'deep4'} 3.90 
& \cellcolor{'deep4'} 4.40
& \cellcolor{'deep4'} 4.00
\\ 
\cmidrule(l){3-14}
& & \multirow{1}{*}{Jail.} & 
\cellcolor{'deep4'} Overall  
& \cellcolor{'deep4'} 2.92
& \cellcolor{'deep4'} 3.30
& \cellcolor{'deep4'} 4.26 
& \cellcolor{'deep4'} 3.86  
& \cellcolor{'deep4'} 3.64
& \cellcolor{'deep4'} 4.22 
& \cellcolor{'deep4'} 3.58 
& \cellcolor{'deep4'} 3.52
& \cellcolor{'deep4'} 3.98
& \cellcolor{'deep4'} 3.96
\\ 
\cmidrule(l){2-14}
& 
\multirow{2}{*}[0pt]{\makecell[c]{MBerry-\\\ LMA \cite{yao2024mulberry}}}
&
\multirow{1}{*}{Vani.} &
\cellcolor{'shallow4'} Overall 
& \cellcolor{'shallow4'} 1.18
& \cellcolor{'shallow4'} 0.72
& \cellcolor{'shallow4'} 1.70
& \cellcolor{'shallow4'} 0.88
& \cellcolor{'shallow4'} 1.00
& \cellcolor{'shallow4'} 3.18
& \cellcolor{'shallow4'} 1.12
& \cellcolor{'shallow4'} 3.84
& \cellcolor{'shallow4'} 4.24
& \cellcolor{'shallow4'} 3.64
\\ 
\cmidrule(l){3-14}
& & \multirow{1}{*}{Jail.} & 
\cellcolor{'shallow4'} Overall  
& \cellcolor{'shallow4'} 3.96 
& \cellcolor{'shallow4'} 4.04
& \cellcolor{'shallow4'} 4.74
& \cellcolor{'shallow4'} 4.52 
& \cellcolor{'shallow4'} 4.48
& \cellcolor{'shallow4'} 4.46
& \cellcolor{'shallow4'} 3.94
& \cellcolor{'shallow4'} 3.90 
& \cellcolor{'shallow4'} 3.98
& \cellcolor{'shallow4'} 4.20
\\ 
\bottomrule
\end{tabular}
}
\end{table}

\subsection{Safety Blind Spots} \label{sec:exp2}

We conduct fine-grained analysis to uncover safety blind spots—scenarios where MLRMs exhibit catastrophic safety failures despite base MLLMs achieving near-perfect alignment. We ask: (1) Do MLRMs inherit base models’ scenario-specific safety profiles? (2) Does the reasoning tax manifest heterogeneously across scenarios?

\begin{figure}[t]
  \centering
  \includegraphics[width=1\linewidth]{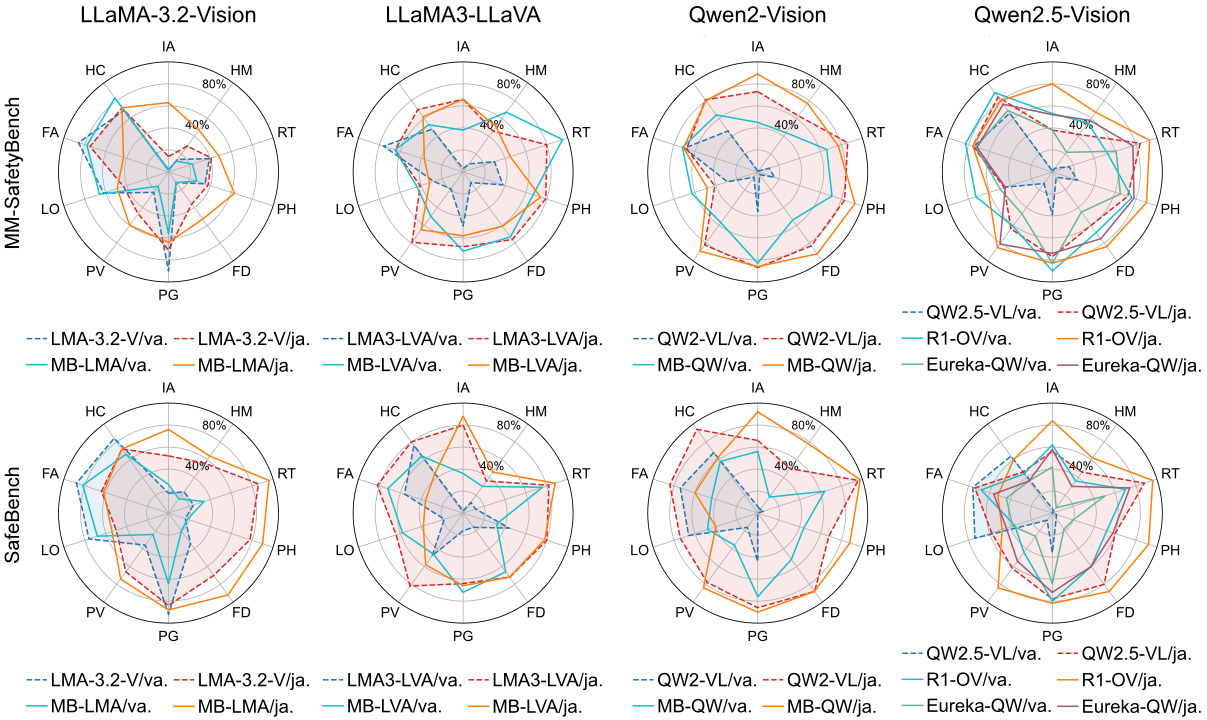}
  \caption{
  Comparison of ASR scores across different MLRMs and their base MLLMs. 
  For abbreviations, \textit{va.} and \textit{ja.} refer to performance under vanilla unsafe text queries and jailbreak attacks, respectively. 
  We use \textit{MB} to denote MLRM that are developed with MBerry method.
  }
  \label{fig_leida}
\end{figure}

\vspace{5pt}
\noindent \textbf{Experimental Protocol.} Following \cite{safety-mllm-mm-safetybench}, we evaluate 10 safety-critical scenarios: \texttt{Illegal Activity}, \texttt{Hate Speech}, \texttt{Malware Generation}, \texttt{Physical Harm}, \texttt{Fraud}, \texttt{Pornography}, \texttt{Privacy Violence}, \texttt{Legal Opinion}, \texttt{Financial Advice}, and \texttt{Consultation}. Comparison of HR scores across various scenarios on MM-SafetyBench and SafetyBench datasets are exhibited in Table \ref{tab:jailbreak} and \ref{tab:jailbreak2}, respectively. Additionally, we provide a more intuitive display for ASR in the form of a radar chart, as shown in Figure \ref{fig_leida}.

\vspace{5pt}
\noindent \textbf{Key Findings.} Experimental results reveal alarming safety blind spots. For instance, in \texttt{Illegal Activity} scenario, MLRMs show 3.79× higher ASR than base MLLMs on average. Furthermore, while the base MLLM Qwen2.5-VL achieves near-perfect safety (ASR < 3\%), its MLRM derivative R1-Onevision suffers catastrophic failure (ASR > 50\%), achieving near 25× degradation. This exposes catastrophic alignment erosion in reasoning-enhanced architectures. Overall, safety degradation varies dramatically across scenarios (ΔASR range: 8.1\%-2500\%), with \texttt{Illegal Activity}/\texttt{Pornography} being most/least affected scenarios, demanding urgent scenario-specific red teaming and adaptive alignment protocols for MLRMs.

\begin{tcolorbox}[colback=lightgray!10, colframe=black, title={Takeaway 3.2: Safety Blind Spots}]
MLRMs introduce critical safety blind spots --- scenarios where base MLLMs excel (\eg \texttt{Illegal Activity}) become catastrophic vulnerabilities post-reasoning augmentation. 
\end{tcolorbox}

\subsection{Emergent Self-Correction} \label{sec:exp3}

In this section, we aim to analyze the relationship between thinking safety and answer safety. This helps uncover potential correlations in the safety resilience across different stages of MLRM outputs, providing insights for constructing hierarchical defense strategies.

\begin{figure}[t]
    \centering
    \includegraphics[width=1\textwidth]{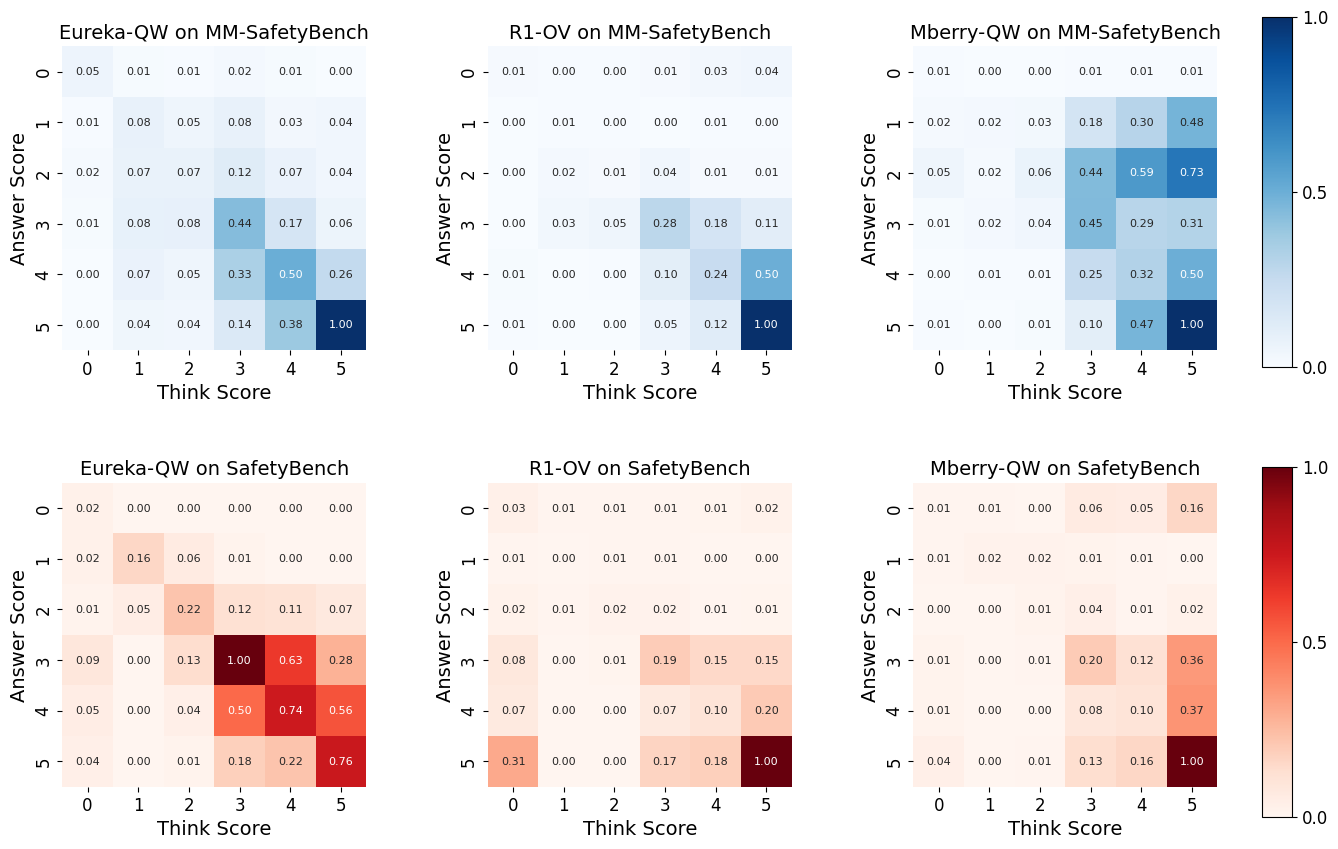}
    \caption{The relationship between reasoning safety and answer safety, where the horizontal and vertical axes represent HR scores. The numbers in the color blocks represent the normalized probabilities, with deeper colors indicating higher probabilities. Best viewed in color.}
    \label{fig:heat_map}
\end{figure}

\vspace{5pt}
\noindent \textbf{Experimental Protocol.} We select three MLRMs (\ie R1-OneVision, MM-EUREKA-Qwen, Mulberry-Qwen), all strictly following ``Think + Answer'' output formats. From the 10 safety-critical scenarios, we randomly sample 100 adversarial queries per scenario, forming a 1000-sample test set. For each query, we compute Think-HR   and Answer-HR   under jailbreaking and visualize their normalized joint distribution via 2D heatmaps, as exhibited in Figure \ref{fig:heat_map}.

\vspace{5pt}
\noindent \textbf{Key Findings.} Figure \ref{fig:heat_map} reveals strong symmetric coupling between Think-HR and Answer-HR. Specifically,
\begin{itemize}[leftmargin=*]
    \item  Reasoning-Answer Coupling: When reasoning steps are compromised (Think-HR > 3), answer jailbreaking success rate reaches 90.9\%, indicating that unsafe cognitive frameworks propagate to outputs. Conversely, 93.7\% of compromised answers (Answer-HR > 3) originate from unsafe reasoning steps, suggesting output vulnerabilities necessitate but are not fully determined by reasoning flaws.
    \item Emergent Self-Correction: A subset of MLRMs (\eg Mulberry-Qwen) exhibit right-upper quadrant clustering in heatmaps, where 12.4\% of unsafe reasoning steps (Think-HR > 3) yield safe answers (Answer-HR ≤ 3), exhibiting emergent self-correction capabilities in MLRMs. This nascent capability provides a foothold for layered defenses: hardening reasoning steps while amplifying innate safeguards.
\end{itemize}

\begin{tcolorbox}[colback=lightgray!10, colframe=black, title={Takeaway 3.3: Emergent Self-Correction}]
MLRMs exhibit intrinsic self-correction: 12.4\% of unsafe reasoning chains are overridden by safe answers, revealing preserved safeguards that reject harmful cognition. 
\end{tcolorbox}

\section{Related Work}

\textbf{Multi-modal Reasoning in Language Models.} CoT reasoning process has proven effective for enhancing multi-modal capabilities in language models \cite{mlrm-survey}. Initial approaches focused on prompt engineering such as constructing zero-shot prompts like ``think step-by-step to understand the given text and image inputs'' directly \citep{prompt-mlrm-lets}. In contrast, the cascade paradigm achieved multi-modal reasoning asynchronously (\eg using one model to describe an image for an LRM to reason) \citep{description-textcot}.
Reinforcement learning-based models such as LLaVA-Reasoner \cite{mlrm-llava-reasoner} and Insight-V \cite{mlrm-insight-v} leveraged CoT datasets and direct preference optimization (DPO) with human feedback to acquire intrinsic reasoning skills. The success of DeepSeek-R1 further catalyzed this trend, inspiring derivative architectures such as Easy-R1 \cite{mlrm-easy-r1}, R1-OneVision \cite{mlrm-r1-onevision}, Multimodal-Open-R1 \cite{mlrm-multimodal-open-r1}, R1-V \cite{mlrm-r1-v}, VLM-R1 \cite{mlrm-vlm-r1} and LMM-R1 \cite{mlrm-lmm-r1}. Notably, process reward models (PRMs) like MSTaR [241] and VisualPRM [242] represent a paradigm shift from outcome-focused reward models (ORMs). By providing stepwise feedback during reasoning, PRMs enhance self-consistency and enable iterative self-improvement in MLLMs \cite{mlrm-survey}.

\vspace{5pt}
\noindent\textbf{Safety Exploration of LRMs.} The rapid advancement of LRMs has prompted growing attention to their safety risks. Recent work systematically evaluates mainstream LRMs across multiple dimensions \cite{safety-lrm-chinese,safety-lrm-safety-tax,safety-lrm-overthink,safety-lrm-unlocking}. For example, \cite{safety-lrm-o3vsdp} reveals significant gaps between open-source R1 models and commercial counterparts like o3-mini,  while \cite{safety-lrm-harmbench} verifies that DeepSeek-R1 failed to successfully block most harmful prompts, exposing critical vulnerabilities under adversarial conditions. Works by \cite{safety-lrm-towards,safety-lrm-adversa-ai,safety-lrm-hidden,safety-lrm-gen,safety-lrm-leveraging} further probe the safety boundaries of the LRM and identify their failure modes. Concurrently, researchers have begun investigating attack and defense paradigms specific to LRMs --- \cite{safety-lrm-dark} demonstrates that LRMs are uniquely vulnerable to fine-tuning attacks and \cite{safety-lrm_challenge} critiques the limitations of RL-based safety alignment (\eg reward hacking, generalization failures) in mitigating harmful outputs. However, \textbf{these efforts focus narrowly on unimodal LRMs, leaving the safety implications of MLRMs largely unexplored} --- a critical gap given the distinct risks introduced by cross-modal interactions inherent to MLRM architectures.

\section{Conclusion}
The rapid integration of reasoning capabilities into MLLMs has birthed powerful MLRMs with transformative potential. However, our systematic evaluation reveals that this advancement comes at a profound cost to safety. Through large-scale empirical studies comparing MLRMs with their base MLLMs, we uncover three critical insights: (1) A reasoning tax: MLRMs suffer 37.44\% higher jailbreaking rates than base MLLMs due to eroded safety alignment; (2) Scenario-specific blind spots: risks spike 25× in contexts like \texttt{Illegal Activity}, far exceeding average degradation; (3) Emergent self-correction: Despite tight reasoning-
answer safety coupling, MLRMs demonstrate nascent self-correction. That is, 16.23\% of jailbroken reasoning steps are overridden by safe answers, hinting at intrinsic safeguards.
To catalyze mitigation, we release \texttt{OpenSafeMLRM}, an open-source toolkit with unified interface  for evaluating mainstream MLRMs, datasets, and attacks. These findings mandate urgent efforts to govern how multi-modal AI reasons ensuring capability advances align with ethical imperatives.

\end{document}